\documentclass[letterpaper, 10 pt, conference]{ieeeconf}  

\IEEEoverridecommandlockouts                              

\usepackage{graphics} 
\usepackage{graphicx}
\usepackage{subfigure}
\usepackage{pifont}
\title{\LARGE \bf
VGG Fine-tuning for Cooking State Recognition
}

\author{Juan Wilches
\thanks{*This work was presented as Project 1 for Deep Learning/Neural Networks course for Spring 2019 at USF}
}

\begin{document}

\maketitle
\thispagestyle{empty}
\pagestyle{empty}

\begin{abstract}

An important task that domestic robots need to achieve is the recognition of states of food ingredients so they can continue their cooking actions. This project focuses on a fine-tuning algorithm for the VGG (Visual Geometry Group) architecture of deep convolutional neural networks (CNN) for object recognition. The algorithm aims to identify eleven different ingredient cooking states for an image dataset. The original VGG model was adjusted and trained to properly classify the food states. The model was initialized with Imagenet weights. Different experiments were carried out in order to find the model parameters that provided the best performance. The accuracy achieved for the validation set was 76.7\% and for the test set 76.6\% after changing several parameters of the VGG model.

\end{abstract}

\section{INTRODUCTION}

The purpose of domestic robotics is to outsource human tasks performed in the households to robotic entities. Some of them are simple like turning on the TV or opening a door. Others like cooking or washing the dishes with their hands are more complicated. Cooking has received special attention in recent years since it is a challenging task that involves several manipulation, learning and recognition stages.

Manipulation is related to the way the robot handles the ingredients. For instance, in order for a robot to cook scrambled eggs, several manipulation techniques need to be used by the robot such as grasping the necessary objects, cracking the eggs, mixing them and then placing them in the stove. Studies of grasping objects by robots are shown in \cite{Lin}, \cite{Lin2015b}, \cite{Lin2015a}.
Works related to learning cooking strategies are discussed in \cite{Paulius2016} and \cite{Paulius2018} where the authors propose a Functional Object Oriented Network (FOON). This network is composed of the knowledge a robot needs to acquire before making an attempt to cook. It shows a relationship among objects, ingredients and the actions that need to be performed in order to properly prepare a dish. The authors in \cite{jelodar2018long} discuss a way to gain knowledge from cooking videos by using computer vision.

Related to the recognition stage convolutional neural networks (CNN) have emerged as a solution to process images and videos for classification. Several architectures have been proposed by different authors like AlexNet \cite{krizhevsky2012imagenet}, Resnet \cite{he2016deep}, GoogleNet \cite{szegedy2015going} and VGG \cite{Simonyan2015vgg}. They have all competed in the ImageNet Challenge \cite{ILSVRC15}. Their architectures are well known classifiers of images for object recognition with high accuracy. Several works have been done for the recognition of cooking ingredient states \cite{jelodar2018resnet}, \cite{salekin2019inception}, \cite{ahmed2018vgg}. These works have focused their goals on fine-tuning algorithms for different architectures. For this project we selected VGG with 19 weight layers in order to perform our own fine-tuning. It has been established that training a convolutional network from scratch is a challenging task \cite{ahmed2018vgg}. Therefore we used the already trained VGG weights for the ImageNet \cite{Deng} dataset.

\section{DATA AND PREPROCESSING}

\begin{figure}[h]
\subfigure[Original]{\includegraphics[width=0.45\columnwidth]{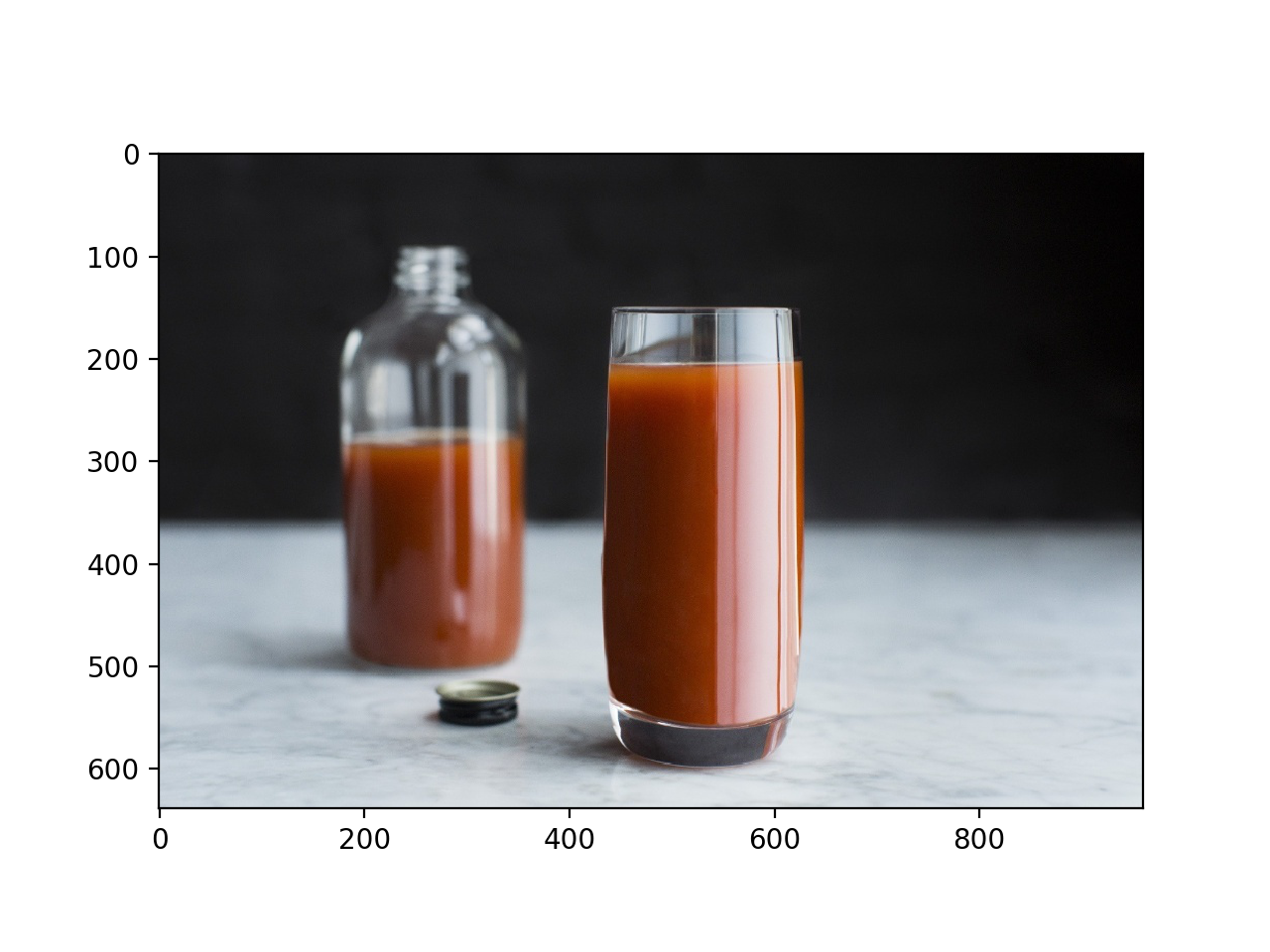}}
\subfigure[Resized]{\includegraphics[width=0.45\columnwidth]{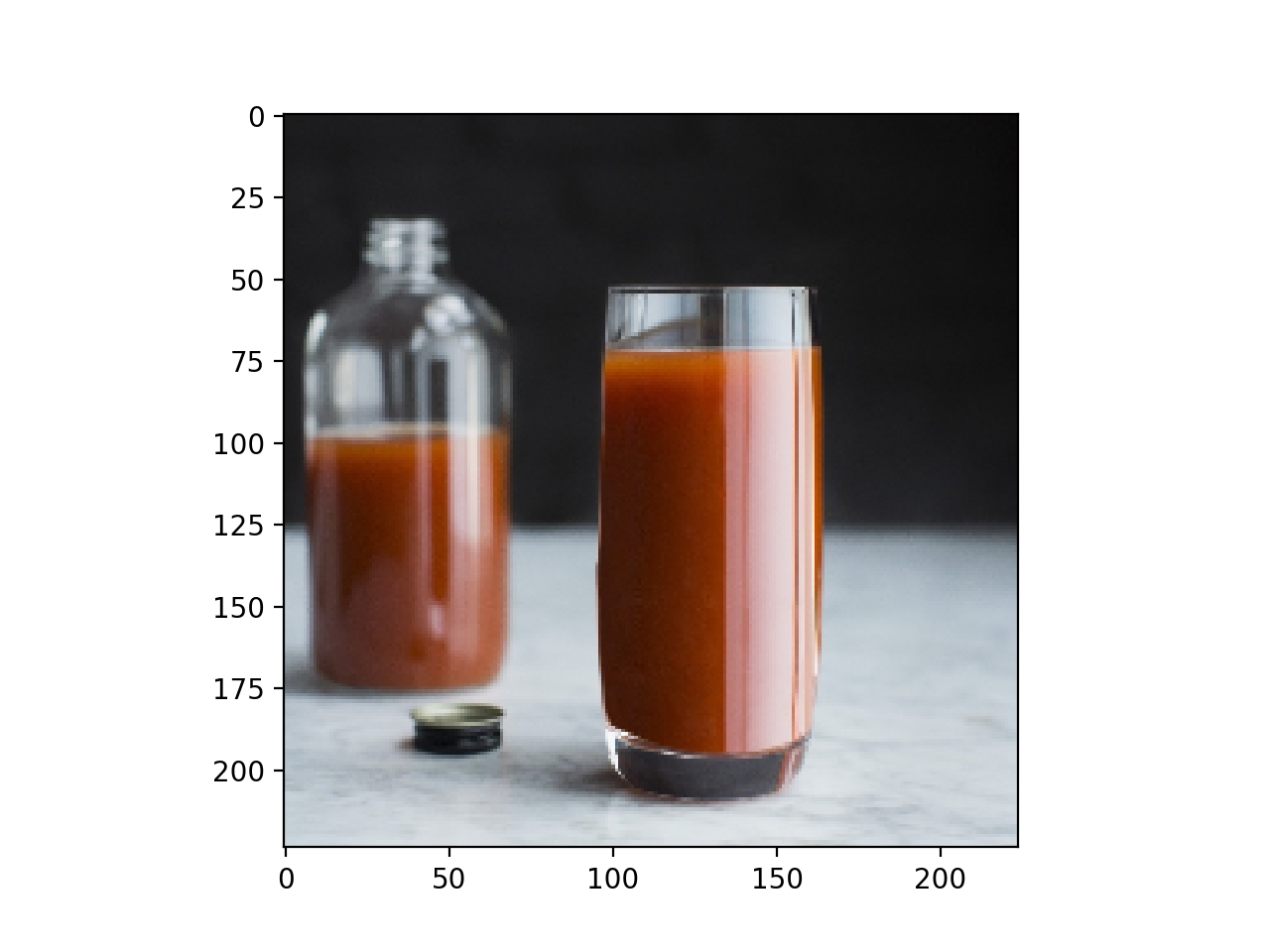}}
\caption{Results of Resize Algorithm}
\label{fig:resize_algorithm}
\end{figure}

The dataset provided has 6,348 images for training, 1,377 for validation and 680 for testing. This is part of the dataset discussed in \cite{jelodar2018identifying}. The amount per class can be seen in table \ref{table_classes}. They are .jpg images with different dimensions. Since the VGG model requires the input images to be RGB of size $224 \times 224$, a resize algorithm was used. The algorithm cropped the images from the center to a size of $224 \times 224$. The result of the algorithm is shown in Fig. \ref{fig:resize_algorithm} for an image of the juiced class. Normalization was also used.

\begin{table}[h]
\caption{Amount of Samples per Class}
\label{table_classes}
\begin{center}
\begin{tabular}{|c||c||c||c|}
\hline
Class & Training & Validation & Test\\
\hline
Floured & 496 & 110 & 57\\
Diced & 511 & 112 & 48\\
Jullienne & 472 & 108 & 56\\
Peeled & 543 & 101 & 42\\
Sliced & 853 & 215 & 103\\
Other & 701 & 143 & 65\\
Grated & 532 & 116 & 68\\
Mixed & 499 & 99 & 55\\
Whole & 745 & 167 & 84\\
Juiced & 491 & 101 & 60\\
Creamy Paste & 505 & 105 & 42\\
\hline
Total & 6,348 & 1,377 & 680\\
\hline
\end{tabular}
\end{center}
\end{table}

\section{METHODOLOGY} \label{methodology}

The original architecture of the VGG model with 19 weights was trained with ImageNet \cite{Deng} dataset, classifies 1,000 classes and has 144MM parameters \cite{Simonyan2015vgg}. As the purpose of the proposed model is to classify 11 classes with a small dataset compared to ImageNet, we decided to remove the two fully connected layers of 4,096 neurons and convert them to only one of 1,024 neurons. Also the fully connected layer with 1,000 neurons that classifies such number of classes was converted to one of 11. These changes led our model to end up with 45MM of parameters to train which is an important reduction from the original model. The proposed modified architecture for fine-tuning is shown in table \ref{table_architecture}.

\begin{table}[h]
\caption{Architecture of the Modified VGG Model}
\label{table_architecture}
\begin{center}
\begin{tabular}{|c|}
\hline
Layers\\
\hline
18 weight layers\\
\hline
input ($224 \times 224$ RGB image)\\
\hline
Conv3-64\\
Conv3-64\\
\hline
Maxpool\\
\hline
Conv3-128\\
Conv3-128\\
\hline
Maxpool\\
\hline
Conv3-256\\
Conv3-256\\
Conv3-256\\
Conv3-256\\
\hline
Maxpool\\
\hline
Conv3-512\\
Conv3-512\\
Conv3-512\\
Conv3-512\\
\hline
Maxpool\\
\hline
Conv3-512\\
Conv3-512\\
Conv3-512\\
Conv3-512\\
\hline
Maxpool\\
\hline
FC-1024\\
\hline
FC-11\\
\hline
Softmax\\
\hline
\end{tabular}
\end{center}
\end{table}

\section{EVALUATION AND RESULTS}

The experiments were carried out using python with tensorflow and GPU NVIDIA GeForce GTX 980 Ti. The batch size was established as 64. We configured our training algorithm for 50 epochs. The learning rate was fixed as 0.0001. We used early stopping when the validation loss started to increase. We used cross-entropy as the loss function for the backpropagation algorithm. At first we tried to train our model using the original VGG architecture and only changing the last layer to have 11 neurons (the amount of classes of our project). The accuracy obtained for this experiment was below 60\%. Then we decided to remove the two fully connected layers as explained in section \ref{methodology}. We trained the modified model with the initial ImageNet weights frozen for the convolutional layers. Hence the weights trained at this stage were the fully connected ones but the accuracy did not increase above 70\%. After this we decided to initialize the model with the ImageNet weights and train it without freezing any weights since we wanted to increase the accuracy. The following experiments were carried out using this last approach.

\subsection{Optimizers}
\begin{figure}[h]
\subfigure[Loss]{\includegraphics[width=0.45\columnwidth]{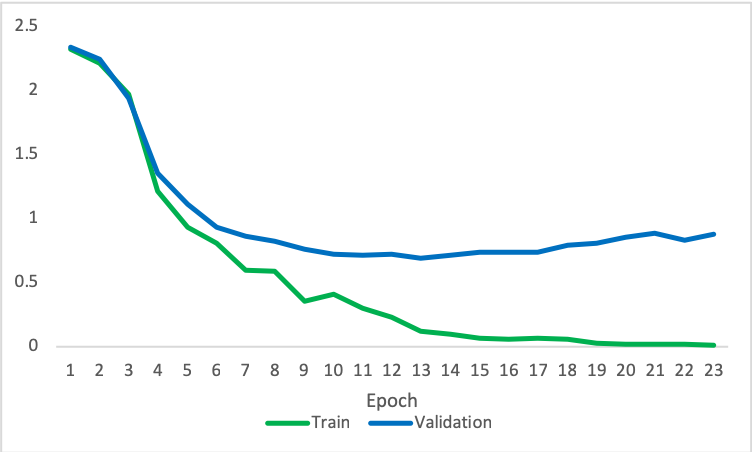}}
\subfigure[Accuracy]{\includegraphics[width=0.45\columnwidth]{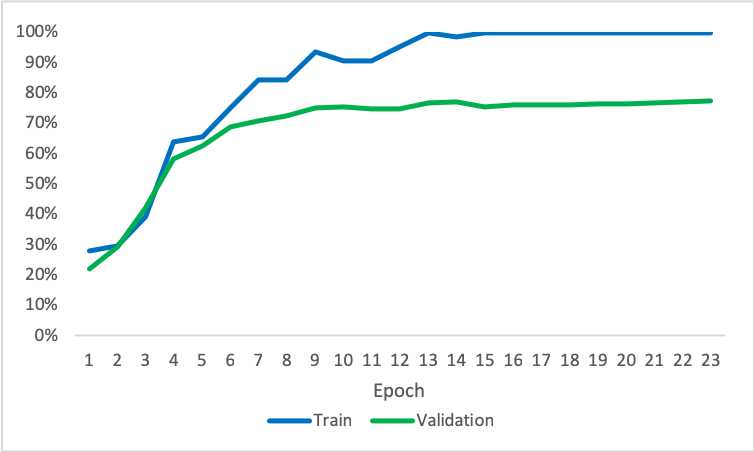}}
\caption{SGD Optimizer}
\label{fig:sgd_optimizer}
\end{figure}

\begin{figure}[h]
\subfigure[Loss]{\includegraphics[width=0.45\columnwidth]{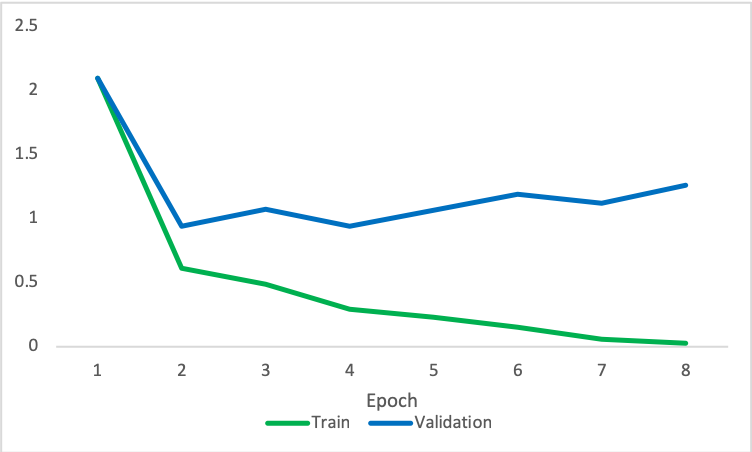}}
\subfigure[Accuracy]{\includegraphics[width=0.45\columnwidth]{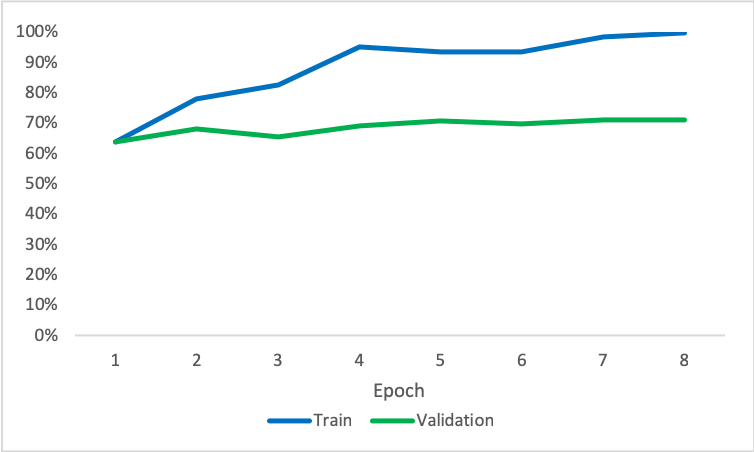}}
\caption{Adam Optimizer}
\label{fig:adam_optimizer}
\end{figure}

\begin{figure}[h]
\subfigure[Loss]{\includegraphics[width=0.45\columnwidth]{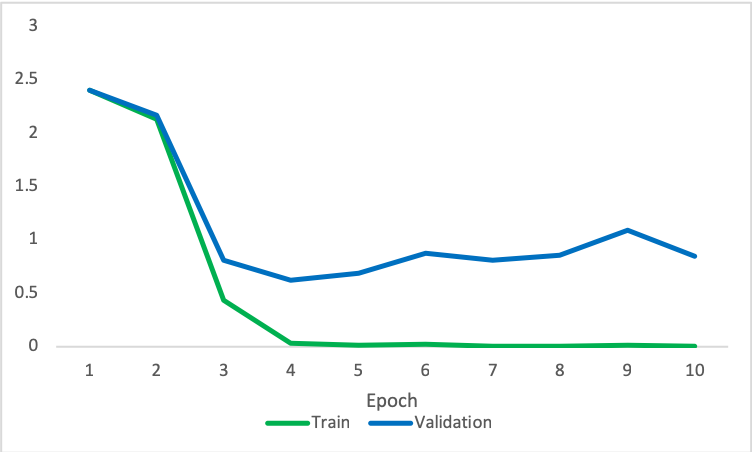}}
\subfigure[Accuracy]{\includegraphics[width=0.45\columnwidth]{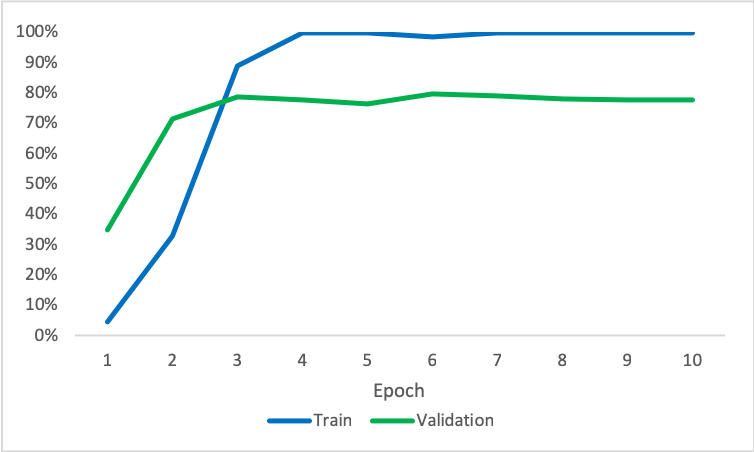}}
\caption{RMSProp Optimizer}
\label{fig:rms_optimizer}
\end{figure}

The most common optimizers used for CNN are Adam and SGD (Stochastic Gradient Descent). We trained our model for those optimizers and also for RMSProp finding that the best accuracy was achieved with SGD. Therefore we chose this optimizer for further experiments. The results of the Loss and Accuracy are shown in Fig. \ref{fig:sgd_optimizer}, \ref{fig:adam_optimizer} and \ref{fig:rms_optimizer} where we can see that after a few epochs the model starts to become overfit. It can be related to the million of parameters that the model updates from each backpropagation algorithm and the few samples of training dataset available. 

\subsection{Number of Neurons}

\begin{figure}[h]
\subfigure[FC-512 Loss and Accuracy ]{\includegraphics[width=0.45\columnwidth]{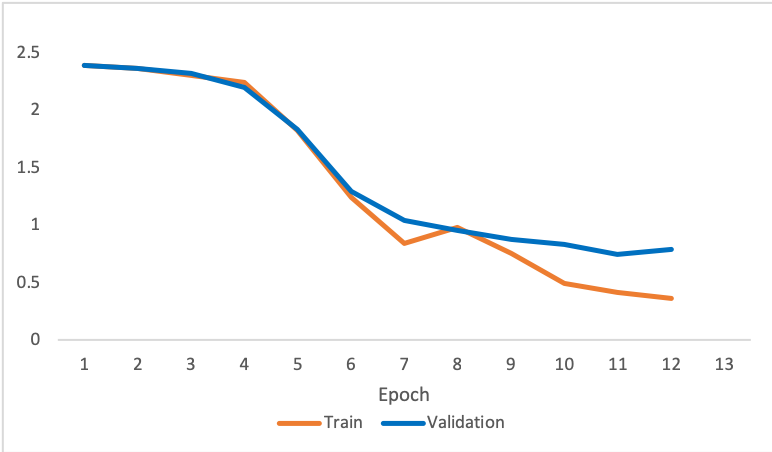} \includegraphics[width=0.45\columnwidth]{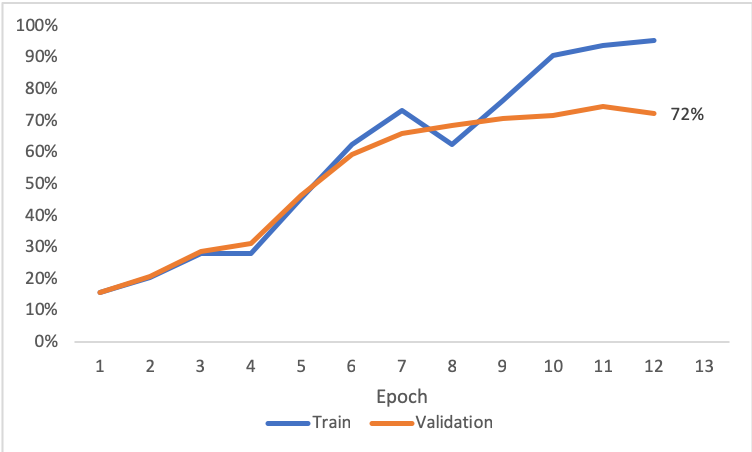}}
\subfigure[FC-1024 Loss and Accuracy \label{fig:1024_neurons}]{\includegraphics[width=0.45\columnwidth]{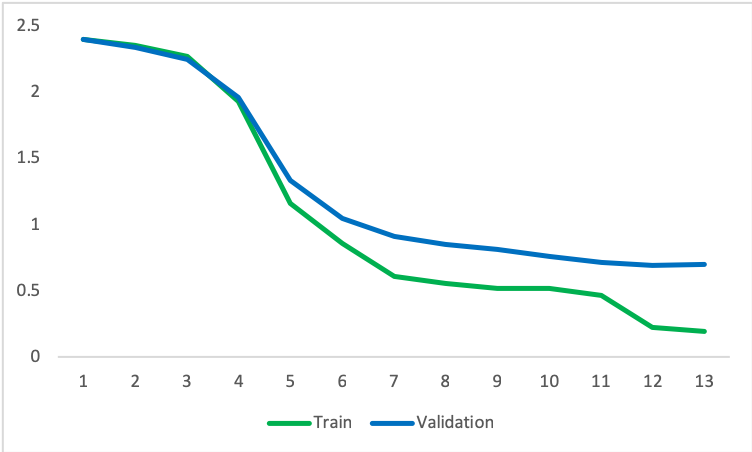} {\includegraphics[width=0.45\columnwidth]{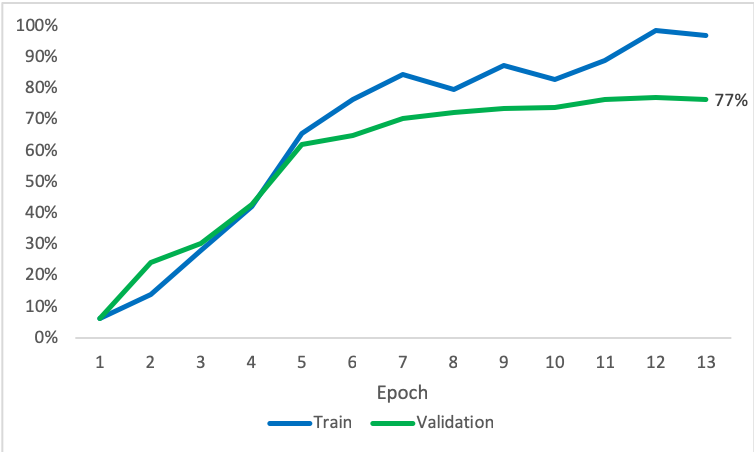}}}
\subfigure[FC-2048 Loss and Accuracy]{\includegraphics[width=0.45\columnwidth]{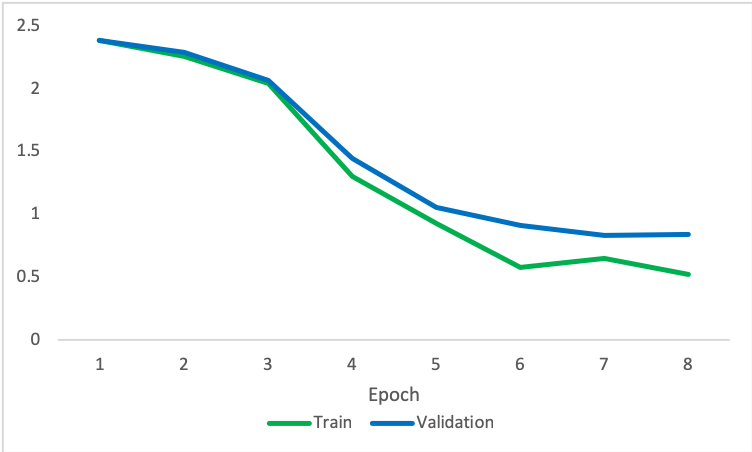} {\includegraphics[width=0.45\columnwidth]{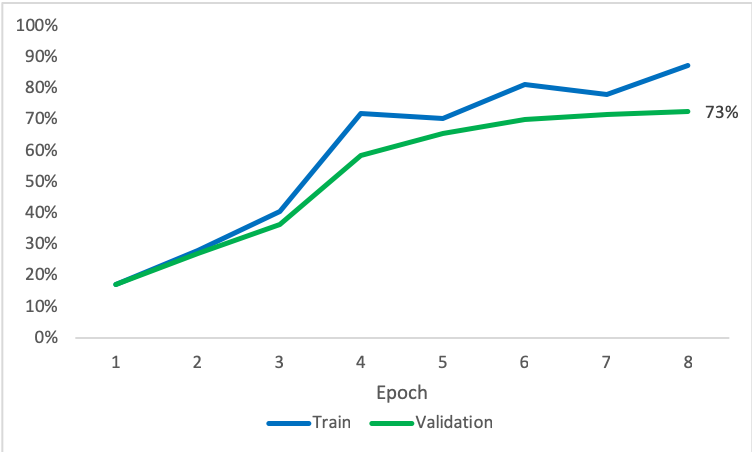}}}
\caption{Change of Fully Connected Layer Number of Neurons}
\label{fig:neurons_change}
\end{figure}

We attempted to improve the accuracy by changing the number of neurons of the fully connected layers. These layers are the ones that perform the classification of the images given the fact that the convolutional layers already extracted the low-level features of the images.  The results can be seen in Fig. \ref{fig:neurons_change}. The highest validation accuracy we were able to achieve was 77\% with one fully connected layer of 1,024 neurons after the convolutional layers. The validation accuracy using 512 neurons was 72\% and for 2,048 neurons was 73\%. The training algorithm was stopped once the validation accuracy started to increase for the three different configurations of neurons. For all three different configurations of neurons we can see that after a few epochs the model becomes overfit.

\subsection{Dropout}

In order to fight overfitting, the model was trained using 50\% and 20\% dropout for the fully connected layer of 1,024 neurons. The accuracy after earlystopping the training algorithm did not go above 74\% for 50\% dropout. It did not increase more than 71\% for 20\% Dropout either. Therefore this regularization technique did not present an improvement in the accuracy of the model. However we can compare these results with Fig. \ref{fig:1024_neurons} and see that the models trained with dropout are less overfit.

\begin{figure}[h]
\subfigure[50\% Dropout Loss and Accuracy ]{\includegraphics[width=0.45\columnwidth]{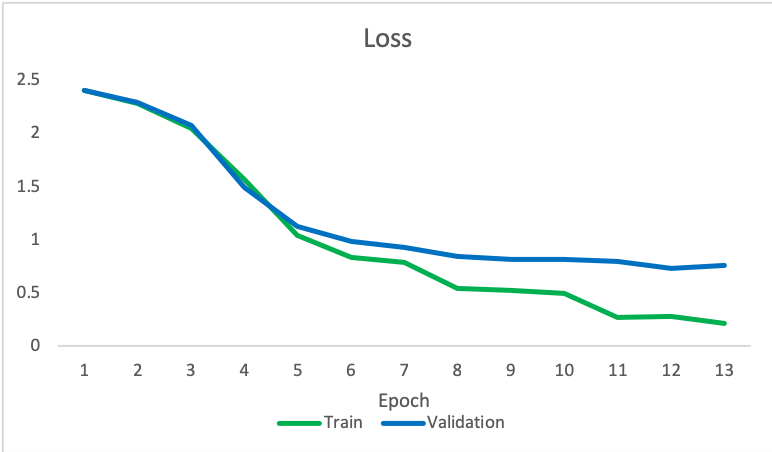} \includegraphics[width=0.45\columnwidth]{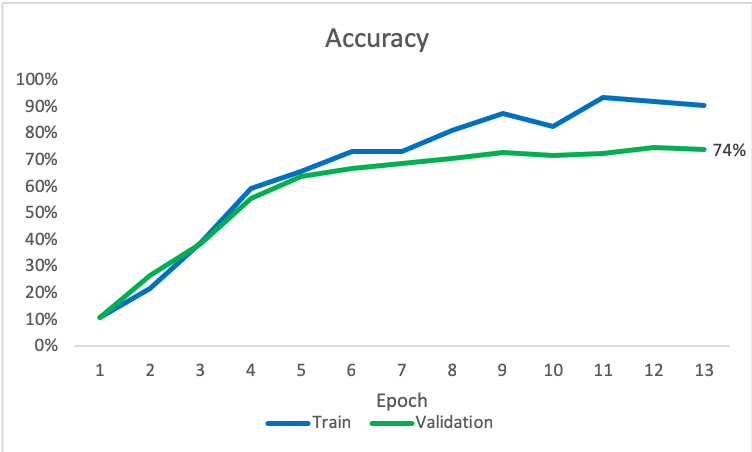}}
\subfigure[20\% Dropout Loss and Accuracy]{\includegraphics[width=0.45\columnwidth]{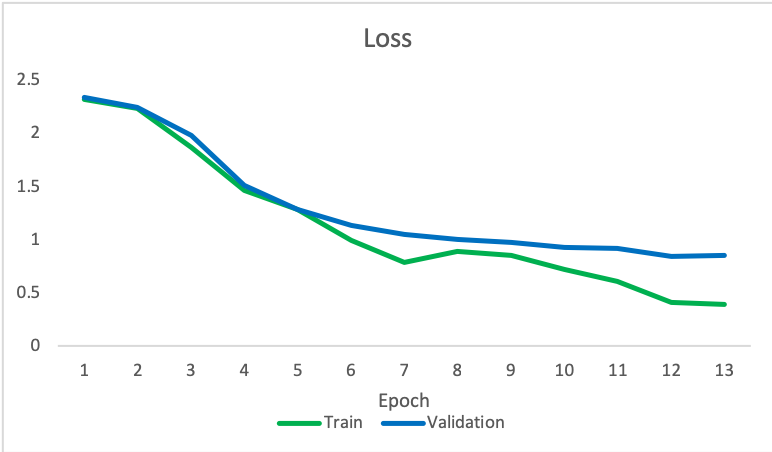} {\includegraphics[width=0.45\columnwidth]{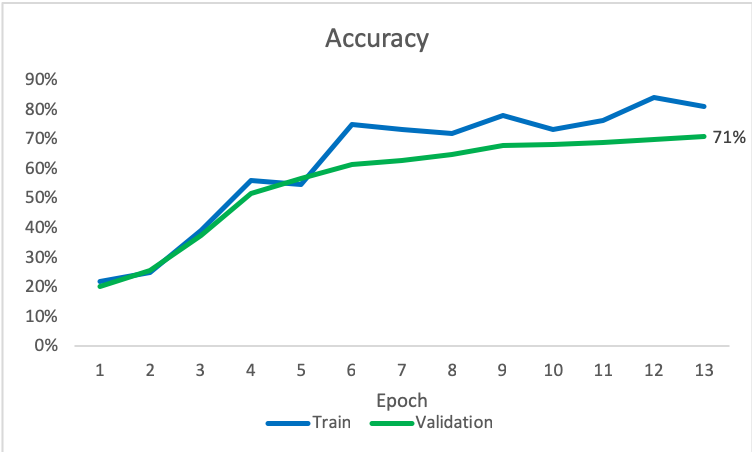}}}
\caption{Training with Dropout of 50\% and 20\%}
\label{fig:dropout}
\end{figure}

\subsection{Testing}
After the conclusion of the different experiments, we selected the model that gave us the highest accuracy. It was the modified VGG model with the loss and accuracy trend shown in Fig. \ref{fig:1024_neurons}. We computed a model confusion matrix for the validation and test datasets in order to verify the correct/incorrect classification of the 11 classes. 

\begin{figure}[h]
\includegraphics[width=\columnwidth]{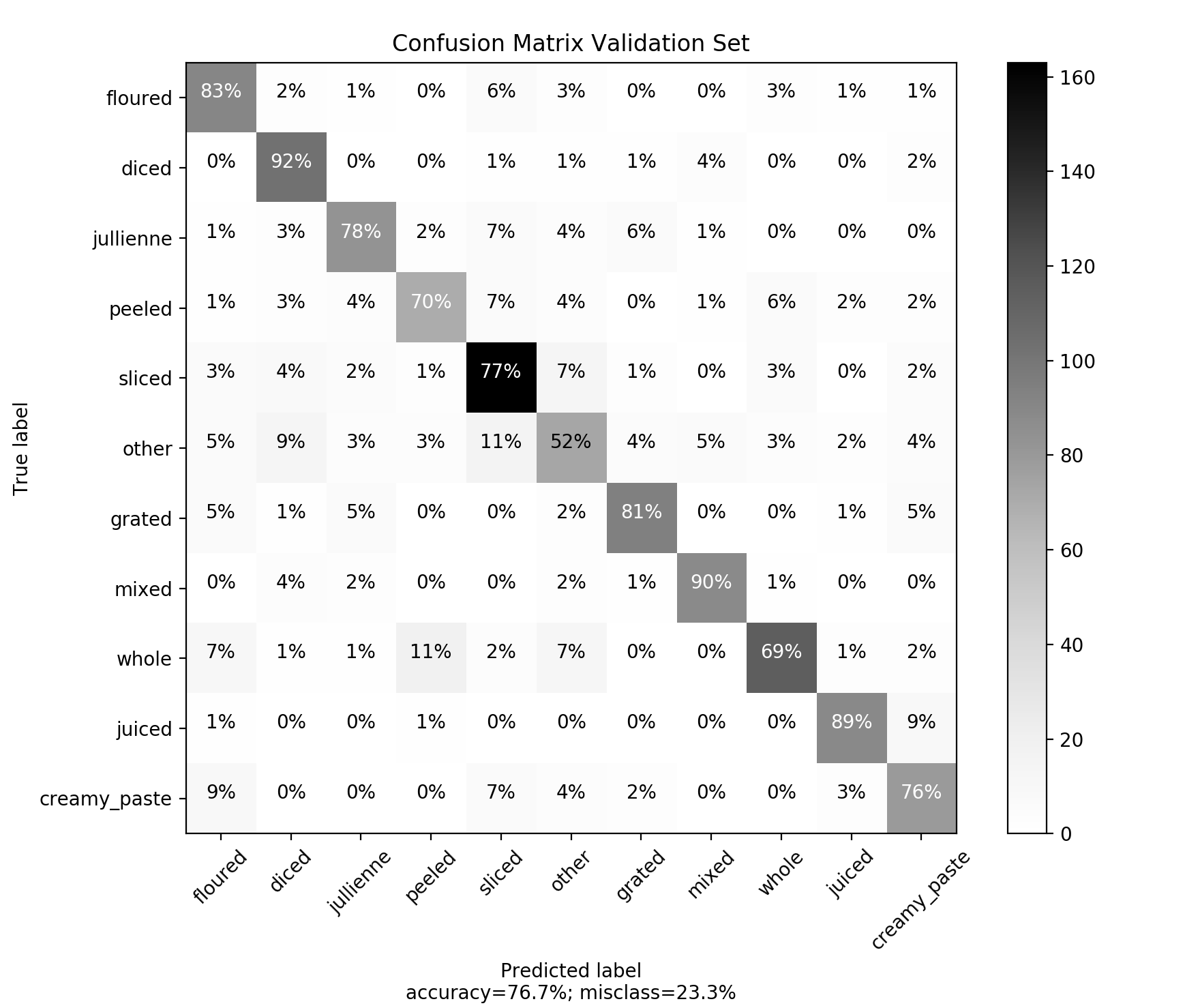}
\caption{Confusion Matrix for the Validation Set}
\label{fig:conf_val}
\end{figure}

\begin{figure}[h]
\subfigure[Other as Sliced]{\includegraphics[width=0.45\columnwidth]{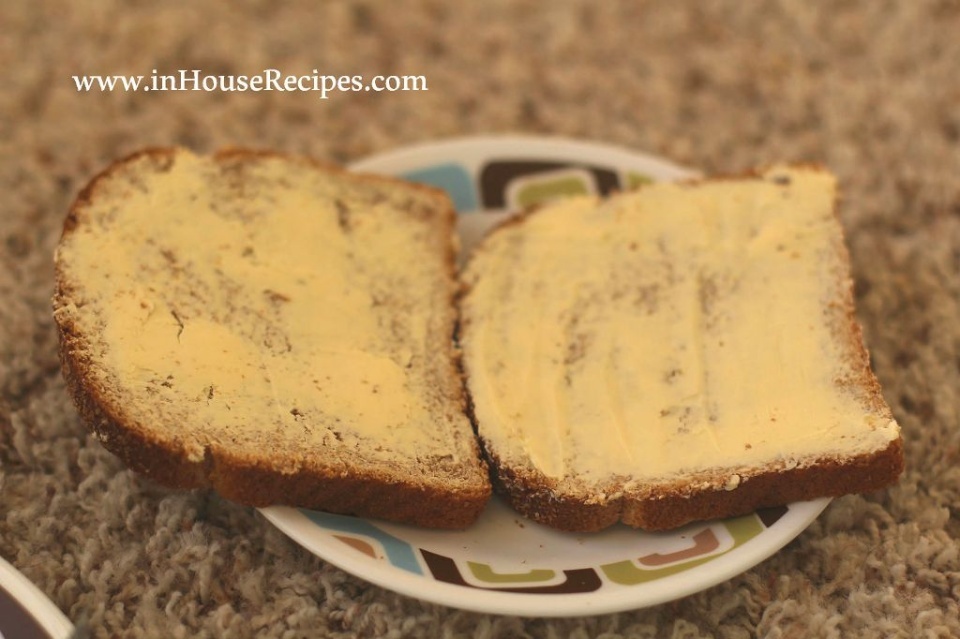}}
\subfigure[Whole as Peeled]{\includegraphics[width=0.45\columnwidth]{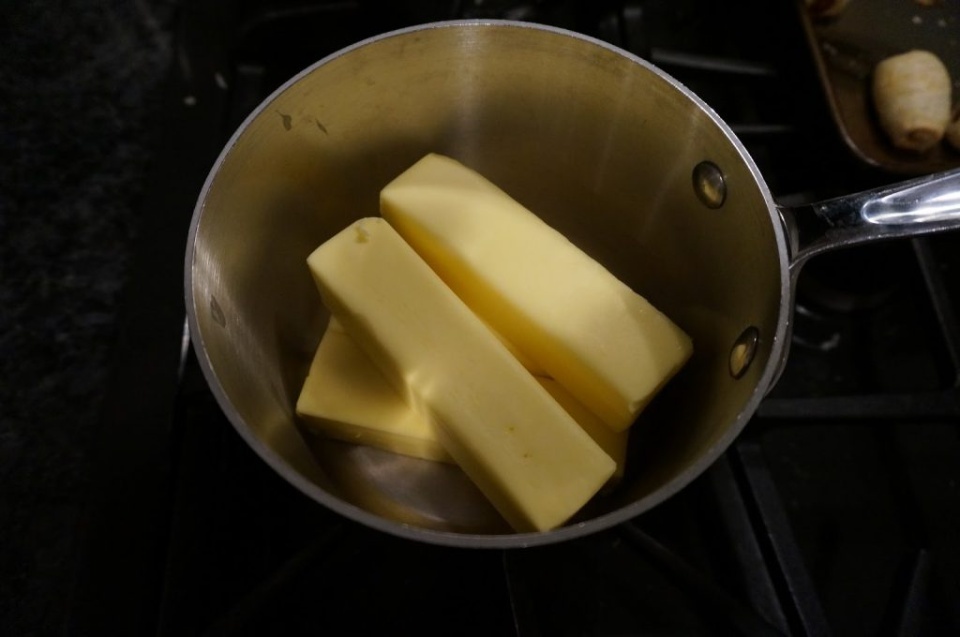}}
\caption{Misclassification for the Validation Set}
\label{fig:val_misclass}
\end{figure}

The confusion matrix for the validation set is shown in Fig. \ref{fig:conf_val}. We can see that the overall accuracy for validation was 76.7\%. The best classified classes were 'Diced' and 'Mixed' with 92\% and 90\% accuracy respectively. Every class was classified with accuracy above 70\% except the class 'Other' that got the least accuracy of 52\%. This can be related to the fact that this is the left out class for images that could not be classified as one of the remaining 10 classes. The model classified 11\% of the samples as sliced and 9\% as diced but at least one sample from the class 'Other' was classified as belonging to one of the remaining 10 classes. We also can see that the model classified 11\% of samples from class 'Whole' as belonging to class 'Peeled'. Examples of the misclassification for the validation set are shown in Fig. \ref{fig:val_misclass} where we see that the neural network tries its best to classify properly the pictures. However even for a human it is complicated to establish the correct state of the shown ingredients.

\begin{figure}[h]
\includegraphics[width=\columnwidth]{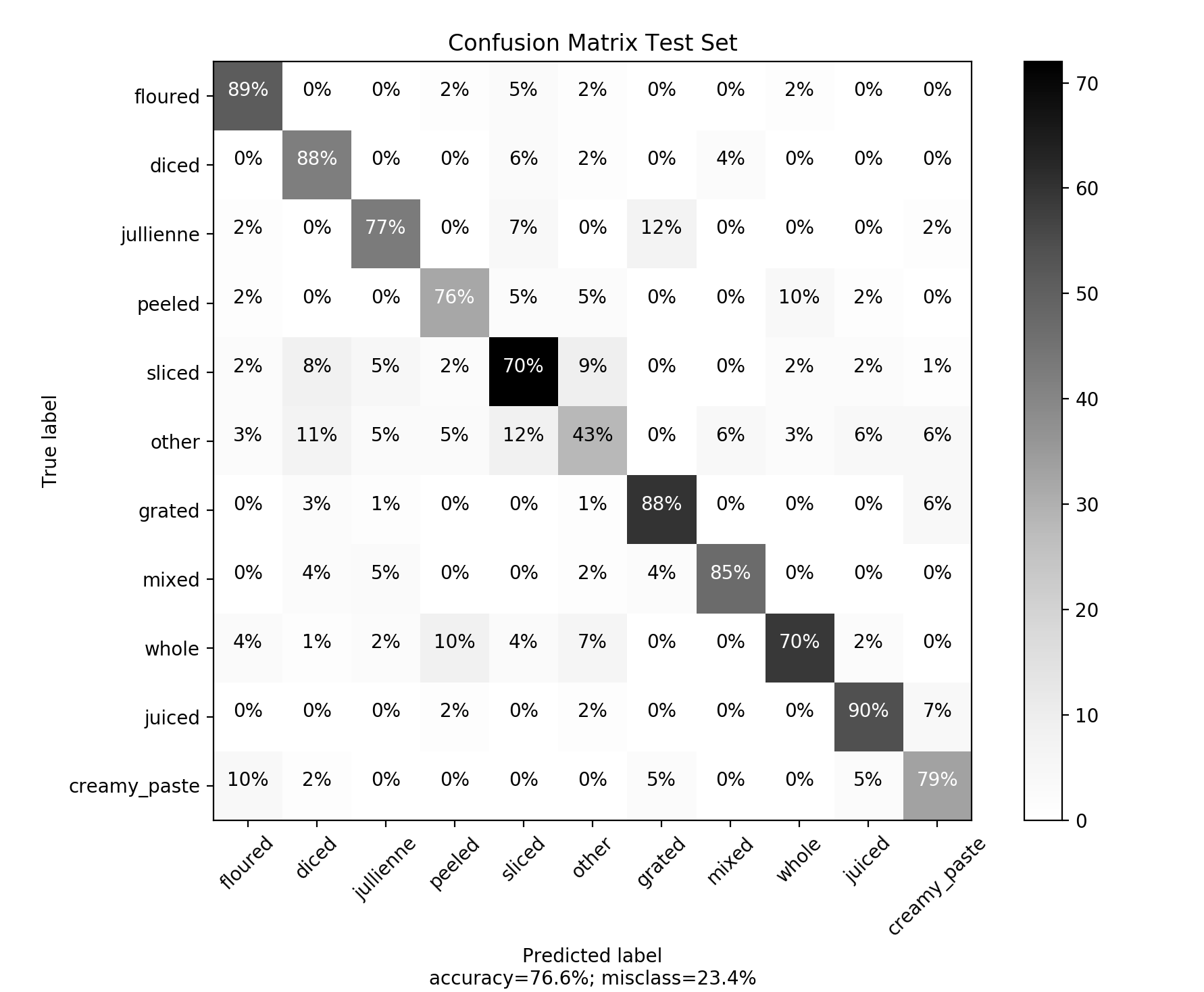}
\caption{Confusion Matrix for the Test Set}
\label{fig:conf_test}
\end{figure}

\begin{figure}[h]
\subfigure[Jullienne as Grated]{\includegraphics[width=0.45\columnwidth, trim={0 180 0 145}, clip]{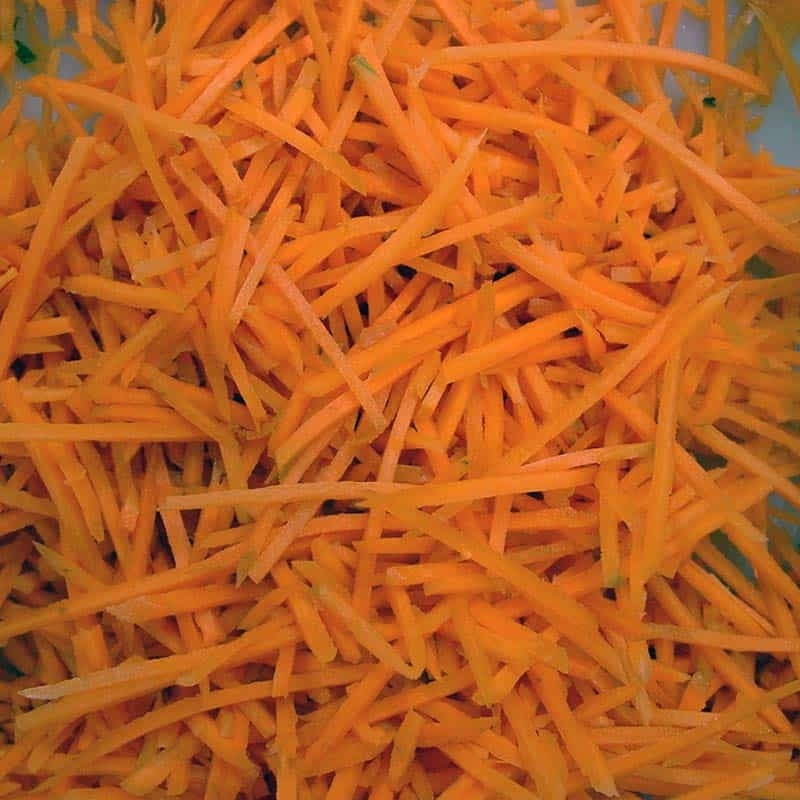}}
\subfigure[Other as Floured]{\includegraphics[width=0.45\columnwidth, trim={50 0 50 0}, clip]{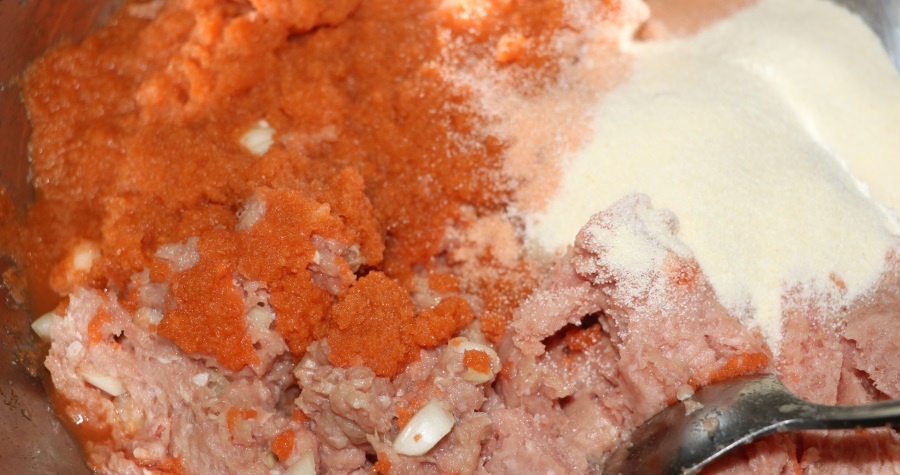}}
\caption{Misclassification for the Test Set}
\label{fig:test_misclass}
\end{figure}

The confusion matrix for the test set is similar to the one computed for the validation set. It is shown in Fig. \ref{fig:conf_test}. We can see that the overall accuracy was 76.6\%. The classes with the highest accuracy for this new matrix are 'Juiced' and 'Floured' with 90\% and 89\% respectively. Again the class with lowest accuracy was 'Other' now with 43\% of correct classification. One particular finding from the test set confusion matrix that the validation's did not show is that the class 'Jullienne' has 12\% of its samples classified as 'Grated'. One example of this misclassification is shown in Fig. \ref{fig:test_misclass}. Again we can see that even for a human it is complicated to discern whether the states that the pictures have according to the dataset are the correct ones.

\section{DISCUSSION}

Ingredient state recognition is an important task that robots need to perform in order to properly achieve the ultimate goal of cooking. In this project we fine-tuned the already well established architecture VGG for object recognition of images. This new model classified 11 different classes of food states. Different experiments were carried out in order to find the modified architecture that best fit the provided dataset. We found that the reduction of parameters of the original VGG model helped to increase the accuracy. Also the initialization of the model with ImageNet weights helped to achieve a high accuracy in a few epochs of training. 

Different experiments related to optimizers, number of neurons and dropout were carried out. The final model accuracy for the validation set was 76.7\% and 76.6\% for test set. Based on the corresponding confusion matrix for each of the sets, we were able to see that there are classes with a high accuracy ~90\% and only the class 'Other' is below 70\%. Further projects should aim to increase this accuracy, enlarge the size of the dataset and verify which other classes are suited for the images contained in the class 'Other'. 

\addtolength{\textheight}{-12cm}   


\bibliographystyle{elsarticle-num}
\bibliography{ref}

\begin{thebibliography}{10}
\expandafter\ifx\csname url\endcsname\relax
  \def\url#1{\texttt{#1}}\fi
\expandafter\ifx\csname urlprefix\endcsname\relax\def\urlprefix{URL }\fi
\expandafter\ifx\csname href\endcsname\relax
  \def\href#1#2{#2} \def\path#1{#1}\fi

\bibitem{Lin}
Y.~Lin, S.~Ren, M.~Clevenger, Y.~Sun, Learning grasping force from
  demonstration, in: IEEE Intl. Conference on Robotics and Automation, 2012,
  pp. 1526--1531.

\bibitem{Lin2015b}
Y.~Lin, Y.~Sun, Grasp planning to maximize task coverage, Intl. Journal of
  Robotics Research 34~(9) (2015) 1195--1210.

\bibitem{Lin2015a}
Y.~Lin, Y.~Sun, Robot grasp planning based on demonstrated grasp strategies,,
  Intl. Journal of Robotics Research 34~(1) (2015) 26--42.

\bibitem{Paulius2016}
D.~Paulius, Y.~Huang, R.~Milton, W.~D. Buchanan, J.~Sam, Y.~Sun, Functional
  object-oriented network for manipulation learning, in: IEEE/RSJ International
  Conference on Intelligent Robots and Systems, IEEE, 2016, pp. 2655--2662.
\newblock \href {http://dx.doi.org/10.1109/IROS.2016.7759413}
  {\path{doi:10.1109/IROS.2016.7759413}}.

\bibitem{Paulius2018}
D.~Paulius, A.~B. Jelodar, Y.~Sun, Functional object-oriented network:
  Construction \& expansion, in: 2018 IEEE International Conference on Robotics
  and Automation (ICRA), IEEE, 2018, pp. 5935--5941.
\newblock \href {http://dx.doi.org/10.1109/ICRA.2018.8460200}
  {\path{doi:10.1109/ICRA.2018.8460200}}.

\bibitem{jelodar2018long}
A.~B. Jelodar, D.~Paulius, Y.~Sun, Long activity video understanding using
  functional object-oriented network, IEEE Transactions on Multimedia.

\bibitem{krizhevsky2012imagenet}
A.~Krizhevsky, I.~Sutskever, G.~E. Hinton, Imagenet classification with deep
  convolutional neural networks, in: Advances in neural information processing
  systems, 2012, pp. 1097--1105.

\bibitem{he2016deep}
K.~He, X.~Zhang, S.~Ren, J.~Sun, Deep residual learning for image recognition,
  in: Proceedings of the IEEE conference on computer vision and pattern
  recognition, 2016, pp. 770--778.

\bibitem{szegedy2015going}
C.~Szegedy, W.~Liu, Y.~Jia, P.~Sermanet, S.~Reed, D.~Anguelov, D.~Erhan,
  V.~Vanhoucke, A.~Rabinovich, Going deeper with convolutions, in: Proceedings
  of the IEEE conference on computer vision and pattern recognition, 2015, pp.
  1--9.

\bibitem{Simonyan2015vgg}
K.~Simonyan, A.~Zisserman, \href{http://arxiv.org/abs/1409.1556}{Very deep
  convolutional networks for large-scale image recognition}, CoRR
  abs/1409.1556.
\newblock \href {http://arxiv.org/abs/1409.1556} {\path{arXiv:1409.1556}}.
\newline\urlprefix\url{http://arxiv.org/abs/1409.1556}

\bibitem{ILSVRC15}
O.~Russakovsky, J.~Deng, H.~Su, J.~Krause, S.~Satheesh, S.~Ma, Z.~Huang,
  A.~Karpathy, A.~Khosla, M.~Bernstein, A.~C. Berg, L.~Fei-Fei, {ImageNet Large
  Scale Visual Recognition Challenge}, International Journal of Computer Vision
  (IJCV) 115~(3) (2015) 211--252.
\newblock \href {http://dx.doi.org/10.1007/s11263-015-0816-y}
  {\path{doi:10.1007/s11263-015-0816-y}}.

\bibitem{jelodar2018resnet}
A.~B. Jelodar, M.~S. Salekin, Y.~Sun,
  \href{http://arxiv.org/abs/1805.06956}{Identifying object states in
  cooking-related images}, CoRR abs/1805.06956.
\newblock \href {http://arxiv.org/abs/1805.06956} {\path{arXiv:1805.06956}}.
\newline\urlprefix\url{http://arxiv.org/abs/1805.06956}

\bibitem{salekin2019inception}
M.~S. Salekin, A.~B. Jelodar, \href{http://arxiv.org/abs/1805.09967}{Cooking
  state recognition from images using inception architecture}, CoRR
  abs/1805.09967.
\newblock \href {http://arxiv.org/abs/1805.09967} {\path{arXiv:1805.09967}}.
\newline\urlprefix\url{http://arxiv.org/abs/1805.09967}

\bibitem{ahmed2018vgg}
K.~B. Ahmed, A.~B. Jelodar, \href{http://arxiv.org/abs/1809.09529}{Fine-tuning
  {VGG} neural network for fine-grained state recognition of food images}, CoRR
  abs/1809.09529.
\newblock \href {http://arxiv.org/abs/1809.09529} {\path{arXiv:1809.09529}}.
\newline\urlprefix\url{http://arxiv.org/abs/1809.09529}

\bibitem{Deng}
J.~Deng, W.~Dong, R.~Socher, L.~Li, K.~Li, L.~Fei-Fei, Imagenet: A large-scale
  hierarchical image database, in: Int. Conf. on Computer Vision and Pattern
  Recognition, 2009, pp. 248--255.

\bibitem{jelodar2018identifying}
A.~B. {Jelodar}, M.~S. {Salekin}, Y.~{Sun}, Identifying object states in
  cooking-related images, arXiv preprint arXiv:1805.06956.

\end{thebibliography}

\end{document}